%% file: paper.tex
\documentclass[a4paper,11pt,table]{article}

\usepackage[utf8]{inputenc}
\usepackage[T1]{fontenc}
\usepackage{lmodern}
\usepackage[margin=22mm]{geometry}
\usepackage{graphicx}
\usepackage{booktabs}
\usepackage{tabularx}
\usepackage{longtable}
\usepackage{array}
\usepackage{xcolor}
\usepackage{hyperref}
\usepackage{fancyhdr}
\usepackage{caption}
\usepackage{subcaption}
\usepackage{float}
\usepackage{amsmath}
\usepackage{amssymb}
\usepackage{microtype}

\definecolor{PrimaryGreen}{HTML}{1B4D3E}
\definecolor{AccentTeal}{HTML}{2E7D6B}
\definecolor{LightTint}{HTML}{F2F7F5}
\definecolor{MutedGray}{HTML}{555555}
\definecolor{RuleGray}{HTML}{CCCCCC}

\hypersetup{
    colorlinks=true,
    linkcolor=PrimaryGreen,
    urlcolor=AccentTeal,
    citecolor=PrimaryGreen,
    pdftitle={Proximal Soil Mapping in Precision Agriculture: Problem Formulation, Grand Architecture, and Open Research Questions},
    pdfauthor={Research Team}
}

\newcolumntype{Y}{>{\raggedright\arraybackslash}X}

\title{\textbf{\color{PrimaryGreen}Proximal Soil Mapping in Precision Agriculture:\\[0.3em]
\Large Problem Formulation, Grand Architecture, and Open Research Questions}}
\author{Yann-M\'eric Sossou \and Kanji Tanaka
\thanks{This work was partially supported by MECSST Grant (23K11270, 26K15002), by KURATA grants and by TATEISI Science And Technology Foundation.}
\thanks{Y.-M. Sossou and K.Tanaka are with Graduate School of Engineering, University of Fukui, Japan.
        {\tt\small tnkknj@u-fukui.ac.jp}}%
}

\date{}

\begin{document}

\maketitle

\begin{abstract}
Acquiring high-resolution, spatially explicit soil information is essential for precision agriculture, yet a persistent measurement gap separates sparse, labor-intensive chemical assays from coarse overhead remote sensing. In this paper, we formalize \textbf{Proximal Soil Mapping} under unstructured outdoor conditions as an integrated scientific problem that couples continuous proximal RGB observations with sparse, asynchronous in-situ sensor measurements to infer spatial soil properties and predictive uncertainty. To realize this vision end-to-end, we propose a five-module system architecture centered on \textbf{SOP Mapping}—a standardized field operating and multimodal synchronization protocol—followed by foundation visual representation, uncertainty-aware prediction, geospatial interpolation, and agronomic decision support. We articulate four open research questions (RQ1--RQ4) governing optical observability limits, semantic artifact disentanglement, spatial generalization, and operational scalability. Finally, we ground this agenda in a prototype field deployment ($N = 242$ synchronized samples; Supplementary Material~A) and a diagnostic sanity check (Supplementary Material~B), verifying non-trivial visual-to-chemical signal recovery for Nitrogen ($\mathrm{OOF}\ R^2 = 0.24$) while demonstrating substantial open research headroom for subsurface macronutrients and spatial mapping.
\end{abstract}

\input m1.tex

\input m2.tex

\input m3.tex


\bibliographystyle{plain}
\bibliography{references}

\clearpage
\appendix
\setcounter{section}{0}
\renewcommand{\thesection}{A.\arabic{section}}
\setcounter{table}{0}
\renewcommand{\thetable}{A.\arabic{table}}
\setcounter{figure}{0}
\renewcommand{\thefigure}{A.\arabic{figure}}

\vspace*{1ex}
{\centering
  {\Large \bfseries\color{PrimaryGreen} Supplementary Material A:\\[0.2em]
  Prototype Experimental System (SOP Mapping) \par}
  \vspace{3ex}
}

\input a1.tex

\input a2.tex

\input a3.tex

\clearpage
\setcounter{section}{0}
\renewcommand{\thesection}{B.\arabic{section}}
\setcounter{table}{0}
\renewcommand{\thetable}{B.\arabic{table}}
\setcounter{figure}{0}
\renewcommand{\thefigure}{B.\arabic{figure}}

\vspace*{1ex}
{\centering
  {\Large \bfseries\color{PrimaryGreen} Supplementary Material B:\\[0.2em]
  Diagnostic Sanity Check and Open Research Headroom \par}
  \vspace{3ex}
}

\input b1.tex

\input b2.tex

\input b3.tex

\end{document}

%% file: m1.tex
\section{Problem Formulation: Proximal Soil Mapping in Precision Agriculture}

Achieving sustainable food production alongside environmental stewardship in precision agriculture requires spatially explicit, high-resolution information regarding soil conditions across agricultural fields \cite{gebbers2010precision, mulla2013twenty}. In operational farmlands, key soil properties—including texture, moisture, availability of macronutrients such as Nitrogen (N), Phosphorus (P), and Potassium (K), pH, electrical conductivity (EC), organic matter content, vegetation cover, and management history—exhibit substantial spatial heterogeneity \cite{cambardella1994field, mcbratney2003digital}. Because such localized variations directly govern crop suitability and yield potential, uniform field-level management based solely on field-average measurements inevitably overlooks local deficiencies or excesses \cite{gebbers2010precision}. Acquiring spatially resolved soil information enables the delineation of relatively homogeneous management zones, scientifically guiding site-specific interventions such as targeted sampling, variable-rate fertilization, precision irrigation, optimal crop allocation, and targeted soil amendment \cite{fraisse2001delineation, mulla2013twenty}.

Conventional soil assessment relies heavily on physical core sampling followed by laboratory wet-chemical analysis or contact-based in-situ sensor measurements \cite{adamchuk2004on, rossEL2016global}. While these direct measurements provide reliable ground-truth reference data, conducting dense sampling across expansive agricultural fields demands prohibitive time, labor, and financial resources \cite{mcbratney2003digital, adamchuk2004on}. Conversely, remote sensing from satellites or unmanned aerial vehicles (UAVs) offers broad spatial coverage but lacks the spatial resolution and sub-canopy visibility needed to capture localized surface soil variations beneath crop cover \cite{mulla2013twenty, maes2019perspectives}. Consequently, there remains a critical measurement gap between sparse, high-cost chemical assays and coarse overhead imagery.

Standard RGB cameras mounted on handheld devices, agricultural machinery, or ground robots offer an accessible, non-destructive modality capable of continuously capturing high-resolution visual observations during routine field traversal \cite{woebbecke1995color, aitkenhead2016digital}. Top-down and proximal RGB imagery contain rich indirect visual cues reflecting underlying soil and crop conditions, including surface coloration, soil texture, surface crusting, visible moisture states, crop residues, and canopy vigor patterns \cite{aitkenhead2016digital, rossEL2016global, gitelson2002novel}. Nevertheless, directly inferring soil chemical properties from uncalibrated outdoor RGB imagery remains highly challenging due to variable ambient illumination, complex canopy-soil occlusion, non-target field artifacts, and the lack of standardized protocols linking continuous visual streams with discrete physical measurements \cite{maes2019perspectives}.

To address this gap, we formalize the problem of \textbf{Proximal Soil Mapping} under realistic, unstructured outdoor conditions: given continuous proximal RGB imagery paired with sparse, asynchronous in-situ sensor measurements collected along operational field trajectories, the objective is to infer spatially resolved soil properties and their associated predictive uncertainty at unobserved locations. Rather than treating soil estimation merely as an isolated regression benchmark on curated images, this formulation treats field data acquisition, cross-modal synchronization, artifact disentanglement, and spatial extrapolation as an integrated scientific problem. In the remainder of this paper, we present a five-module grand architecture to realize this vision (Section~2) and articulate the core open research questions that must be resolved to make proximal soil mapping operationally trustworthy (Section~3), supported by an initial experimental prototype (Supplementary Material~A) and a diagnostic sanity-check study (Supplementary Material~B).

%% file: m2.tex
\section{Grand Vision: Five-Module Architecture for AI-Driven Soil Mapping and Decision Support}

To move beyond isolated, static soil estimation toward an end-to-end operational system, we envision an integrated precision agriculture framework that seamlessly links physical field acquisition with actionable agronomic decision-making \cite{gebbers2010precision, mcbratney2003digital}. Rather than treating computer vision, soil chemistry modeling, and geostatistics as disconnected tasks, our overarching system architecture is organized into five interconnected modules: (1) field data acquisition via SOP Mapping, (2) image preparation and visual feature extraction, (3) uncertainty-aware soil-property prediction, (4) geospatial integration and management-zone mapping, and (5) agricultural decision support. Together, these modules define a scalable pipeline capable of transforming raw, unstructured field traversals into site-specific management prescriptions.

\textbf{Module 1: Field Data Acquisition via SOP Mapping.} The foundation of any trustworthy soil-mapping system is reproducible, spatially anchored ground data \cite{adamchuk2004on}. In practical agricultural settings—where high-precision real-time kinematic (RTK) GNSS may be unavailable, intermittent, or cost-prohibitive—ad-hoc manual sampling easily suffers from asynchronous sensor clocks, non-contact transit noise, and ambiguous spatial provenance. To overcome these bottlenecks, Module~1 introduces \textbf{SOP Mapping}, which formalizes standardized field traversal protocols (such as partitioned strip layouts, U-shaped walking paths, fixed stationary probe-stabilization intervals, and dual-perspective close-up/wide-angle video auditing) directly into the spatial data-acquisition pipeline. By coupling hardware-grounded filtering rules with cross-modal timestamp synchronization, SOP Mapping converts routine human field operations into auditable, spatially structured multimodal datasets.

\textbf{Module 2: Image Preparation and Visual Feature Extraction.} Proximal RGB imagery captured in open fields inevitably contains non-target visual artifacts—such as harsh cast shadows, operator footwear, sensor probes, and equipment enclosures—alongside complex mixtures of exposed soil, living canopy, and crop residues \cite{woebbecke1995color, aitkenhead2016digital}. Module~2 addresses this challenge by separating scene components prior to representation learning. Specifically, foundation segmentation models such as Segment Anything Model 2.1 (SAM~2.1) \cite{ravi2024sam2} enable semantic partitioning of exposed soil, vegetation, and residues from artificial occlusions, while self-supervised visual encoders such as DINOv3 \cite{oquab2024dinov2} extract generalizable, high-dimensional descriptors capturing surface color, soil clod texture, moisture cues, and canopy structure.

\textbf{Module 3: Uncertainty-Aware Soil-Property Prediction.} Scientifically, it cannot be assumed \textit{a priori} that all chemical soil variables are equally observable from surface RGB imagery alone \cite{rossEL2016global}. Module~3 maps the extracted visual representations to target soil properties (e.g., Nitrogen, Phosphorus, Potassium, moisture, and organic matter) while explicitly quantifying predictive uncertainty \cite{gal2016dropout, kendall2017uncertainties}. Incorporating uncertainty estimation allows the system to distinguish between visually well-constrained variables (such as canopy-linked Nitrogen status) and weakly observable subsurface properties, flagging low-confidence predictions or triggering targeted physical follow-up sampling rather than propagating overconfident estimates downstream \cite{kendall2017uncertainties}.

\textbf{Modules 4 and 5: Geospatial Integration and Agricultural Decision Support.} Finally, Modules~4 and~5 translate point-wise predictions into field-wide operational decisions. In Module~4, validated soil estimates and their associated uncertainty layers are fused with GNSS coordinates or SOP-derived spatial anchors and interpolated via geostatistical methods (such as kriging \cite{matheron1963principles, cressie1993statistics}) to generate continuous spatial property maps and delineate homogeneous management zones \cite{mcbratney2003digital, fraisse2001delineation}. In Module~5, these spatial soil layers are combined with crop-specific physiological requirements, climatic and weather data, water availability, topography, management history, and economic or environmental constraints to recommend suitable crops, variable-rate fertilization and irrigation schedules, and targeted soil-amendment strategies (e.g., adjusting pH, nutrient balance, or water-holding capacity) \cite{gebbers2010precision, zhai2020decision}.

%% file: m3.tex
\section{Open Research Questions}

Realizing the five-module architecture envisioned in Section~2 under operational agricultural conditions requires resolving fundamental scientific and methodological bottlenecks that precede routine model tuning. Simply benchmarking off-the-shelf regressors on small, curated image sets cannot establish whether proximal visual sensing can be trusted across diverse fields, seasons, and soil types. To guide systematic progress in this emerging domain, we formulate four core open research questions (RQ1--RQ4) spanning optical physics, computer vision, statistical learning, and field engineering.

\textbf{RQ1 (Optical Observability Limits): Which soil chemical properties can be reliably inferred from proximal RGB observations, at what spatial scale, and under what environmental conditions?} Surface RGB imagery primarily captures visible reflectance from the uppermost soil layer and overlying vegetation canopy \cite{rossEL2016global}. While macronutrients such as Nitrogen (N) often manifest indirectly through canopy vigor and leaf greenness \cite{gitelson2002novel}, subsurface properties such as Phosphorus (P), Potassium (K), pH, and cation exchange capacity may exhibit weak or indirect optical coupling \cite{rossEL2016global}. Establishing a trustworthy soil-mapping framework requires rigorously mapping the physical boundary between visually observable variables and properties that demand direct contact sensing or multi-modal fusion.

\textbf{RQ2 (Artifact and Semantic Disentanglement): How can true agronomic signals from exposed soil and living vegetation be semantically disentangled from outdoor illumination variability and non-target field artifacts?} Unstructured ground-level field imagery frequently contains severe visual confounders, including harsh cast shadows, specular soil glare, operator footwear, sensor probes, equipment enclosures, and senescent crop residues \cite{woebbecke1995color}. Simple color-space thresholding can suppress high-contrast artificial objects, but it fails to semantically separate bare soil clods from dead plant litter or complex canopy backgrounds \cite{woebbecke1995color}. Investigating how foundation segmentation models (e.g., SAM~2.1 \cite{ravi2024sam2}) and self-supervised visual representations (e.g., DINOv3 \cite{oquab2024dinov2}) perform on small, noisy agronomic datasets without overfitting remains a critical open challenge.

\textbf{RQ3 (Spatial Generalization and Uncertainty Quantification): How can predictive models avoid spatial-temporal leakage and provide calibrated uncertainty estimates under small sample sizes ($N < 300$) and strong spatial autocorrelation?} Because reference wet-chemical assays and in-situ probe measurements are labor-intensive to acquire, operational field datasets typically suffer from limited sample counts alongside high predictor collinearity. Evaluating models via naive random cross-validation risks inflating accuracy due to spatial proximity between training and test spots \cite{roberts2017cross}. Developing spatially and temporally grouped validation protocols \cite{roberts2017cross} alongside principled uncertainty quantification \cite{kendall2017uncertainties}—capable of abstaining or triggering targeted physical sampling in high-uncertainty zones—is essential for safe downstream decision-making.

\textbf{RQ4 (SOP-to-Spatial Scalability): How can manual field Standard Operating Procedures (SOPs) and asynchronous multi-sensor logs be formalized into scalable, reproducible spatial data-acquisition pipelines?} In real-world farm deployments, high-precision RTK-GNSS receivers may be unavailable or degraded, while handheld cameras and contact soil probes operate on independent clocks with distinct stabilization dynamics. Codifying human traversal rules, hardware-grounded signal filtering, and multi-perspective visual auditing into a standardized \textbf{SOP Mapping} protocol—and validating its portability across different fields, operators, and survey dates—remains a foundational systems challenge.

\textbf{Scope of This Paper and Empirical Grounding.} By formalizing the proximal soil-mapping problem (Section~1), defining a five-module grand architecture (Section~2), and articulating RQ1--RQ4 above, this paper establishes a unified research agenda for vision-based soil mapping. To ground these concepts in concrete field practice, \textbf{Supplementary Material~A} details our initial prototype implementation of Module~1 (\textbf{SOP Mapping}), documenting the portable sensor-box hardware, four-person traversal protocol, automatic non-contact filtering, and cross-modal synchronization ($\tau_{\max} = 84\text{ s}$) deployed to construct a synchronized dataset ($N = 242$). Subsequently, \textbf{Supplementary Material~B} presents a lightweight diagnostic \textbf{sanity check} on this prototype dataset, verifying that non-trivial agronomic signal is preserved (e.g., for Nitrogen) while empirically demonstrating that weak subsurface observability (for Phosphorus and Potassium), heuristic segmentation bottlenecks, and spatial generalization remain largely open problems requiring sustained investigation.

%% file: a1.tex
\section{Portable Sensing System and Multimodal Capture Architecture}

\begin{figure}[H]
    \centering
    \includegraphics[width=\textwidth]{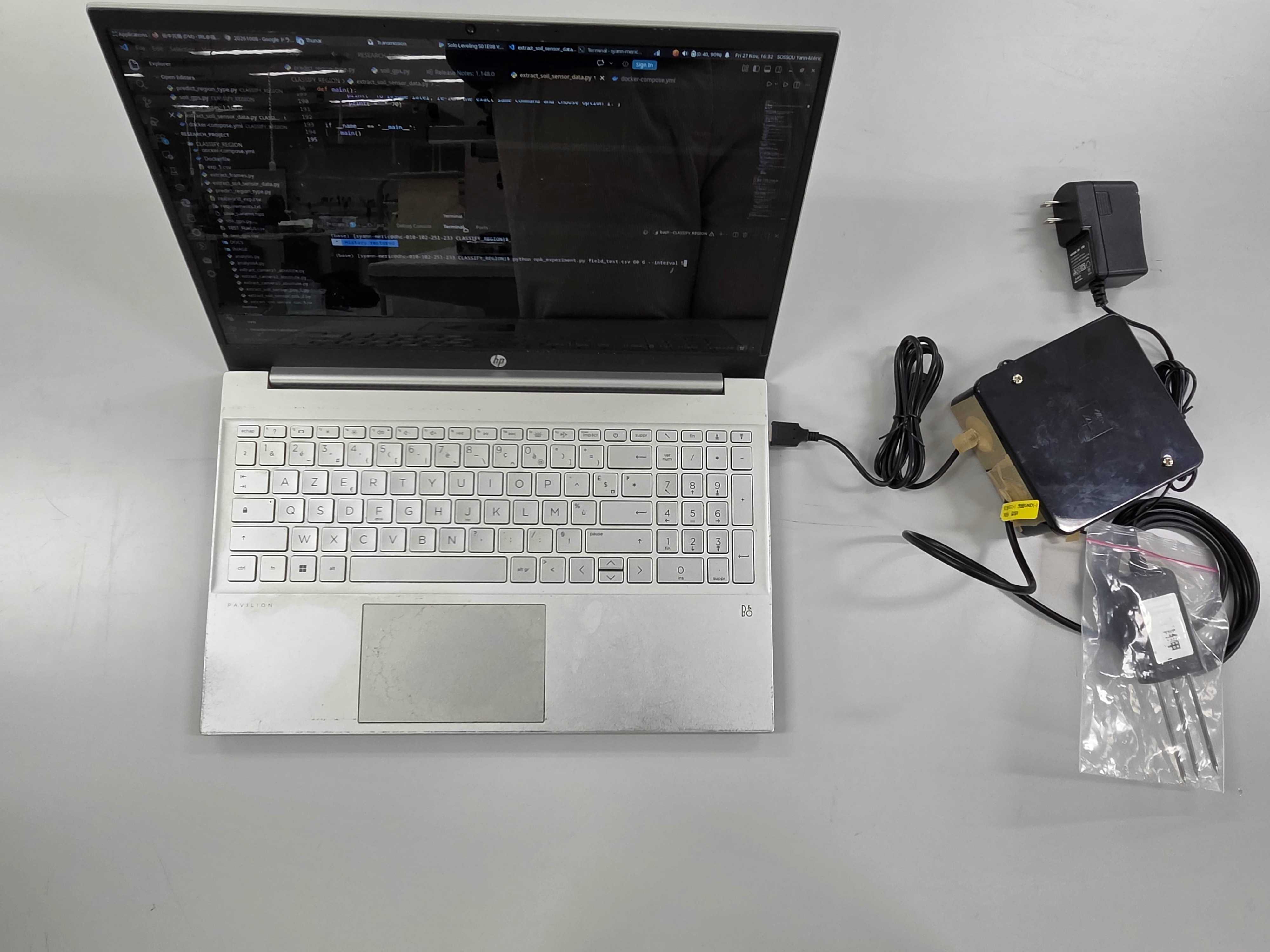}
    \caption{Assembly stages and benchtop integration of the portable RS485 soil sensor box.}
    \label{fig:assembly}
\end{figure}

To instantiate Module~1 (\textbf{SOP Mapping}) in real agricultural terrain without relying on bulky stationary infrastructure, we developed a self-contained, portable soil sensing system centered on an RS485 Modbus RTU macronutrient probe (measuring Nitrogen, Phosphorus, and Potassium in $\text{mg/kg}$ at 9600\,baud). As detailed in Table~\ref{tab:equipment} and Figure~\ref{fig:assembly}, the probe's differential signal lines (\texttt{A+}, \texttt{B-}) and power leads (\texttt{VCC}, \texttt{GND}) are integrated inside a dust- and shock-resistant protective junction box (\texttt{Mirai Industry PVK-ANK}) via an industrial Waveshare USB-to-RS485 converter and a screw-terminal DC barrel jack adapter. Regulated 12\,V DC excitation power is supplied off-grid by a portable AC/DC power station (Jackery Portable Power 400, 293\,Wh) coupled with a 12\,V DC switching adapter, while a dedicated Python acquisition script (\texttt{extract\_soil\_sensor\_data.py} / \texttt{npk\_experiment.py}) running on a host laptop continuously polls the sensor holding registers (starting at \texttt{0x001E}) at a nominal 2\,s interval and appends timestamped records to CSV.

\begin{table}[H]
\centering
\caption{Hardware components used to construct the portable RS485 soil sensor-box system.}
\label{tab:equipment}
\renewcommand{\arraystretch}{1.35}
\begin{tabularx}{\textwidth}{>{\bfseries}p{3.5cm} Y p{3.6cm} c}
\toprule
\rowcolor{LightTint}
Component & Function in the System & Model / Specification & Image \\
\midrule
12\,V DC power supply &
Provides regulated 12\,V DC excitation power required by the RS485 soil probe circuit. &
12\,V DC switching adapter (Model: B00IGU4VY2) &
\raisebox{-0.45\height}{\includegraphics[width=2.0cm]{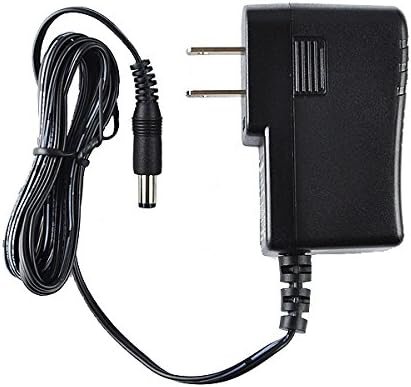}} \\
\midrule
DC jack adapter &
Provides a screw-terminal DC barrel interface for secure internal power distribution. &
Terminal-block DC jack (Akizuki Denshi, Cat.\ No.\ 108849) &
\raisebox{-0.45\height}{\includegraphics[width=2.0cm]{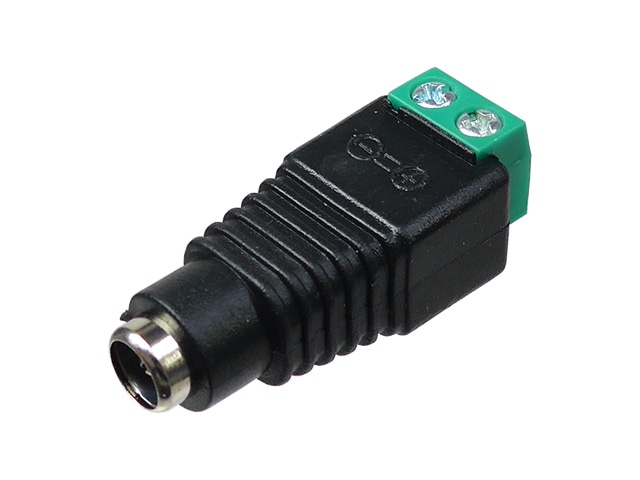}} \\
\midrule
USB-to-RS485 converter &
Provides isolated serial communication between the RS485 Modbus probe and the host laptop over USB. &
Waveshare Industrial USB-to-RS485 Converter &
\raisebox{-0.45\height}{\includegraphics[width=2.0cm]{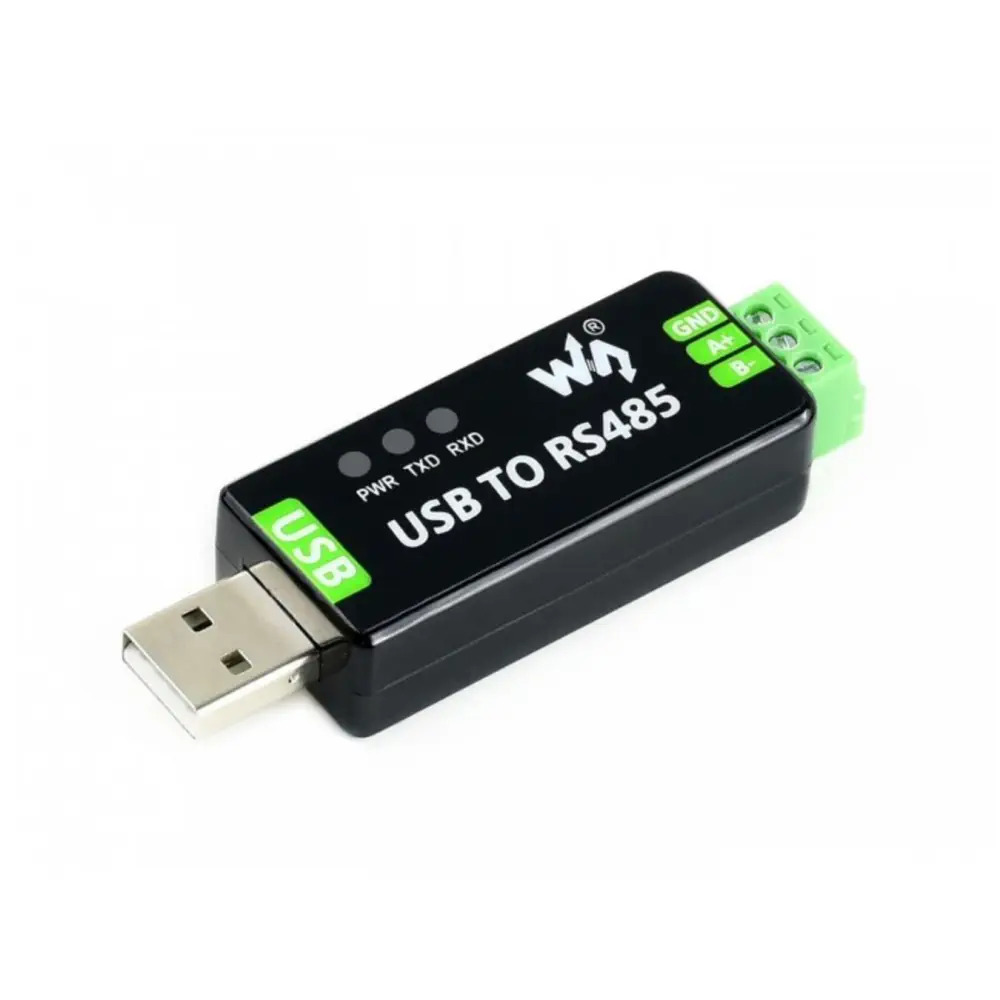}} \\
\midrule
USB cable &
Connects the USB-to-RS485 converter to the host laptop for serial data transfer. &
Standard USB-A cable (supplied with converter) &
\textit{Included} \\
\midrule
Portable power station &
Supplies off-grid AC/DC electrical power to the 12\,V DC adapter during field traversal. &
Jackery Portable Power 400 (293\,Wh / 200\,W) &
\raisebox{-0.45\height}{\includegraphics[width=2.0cm]{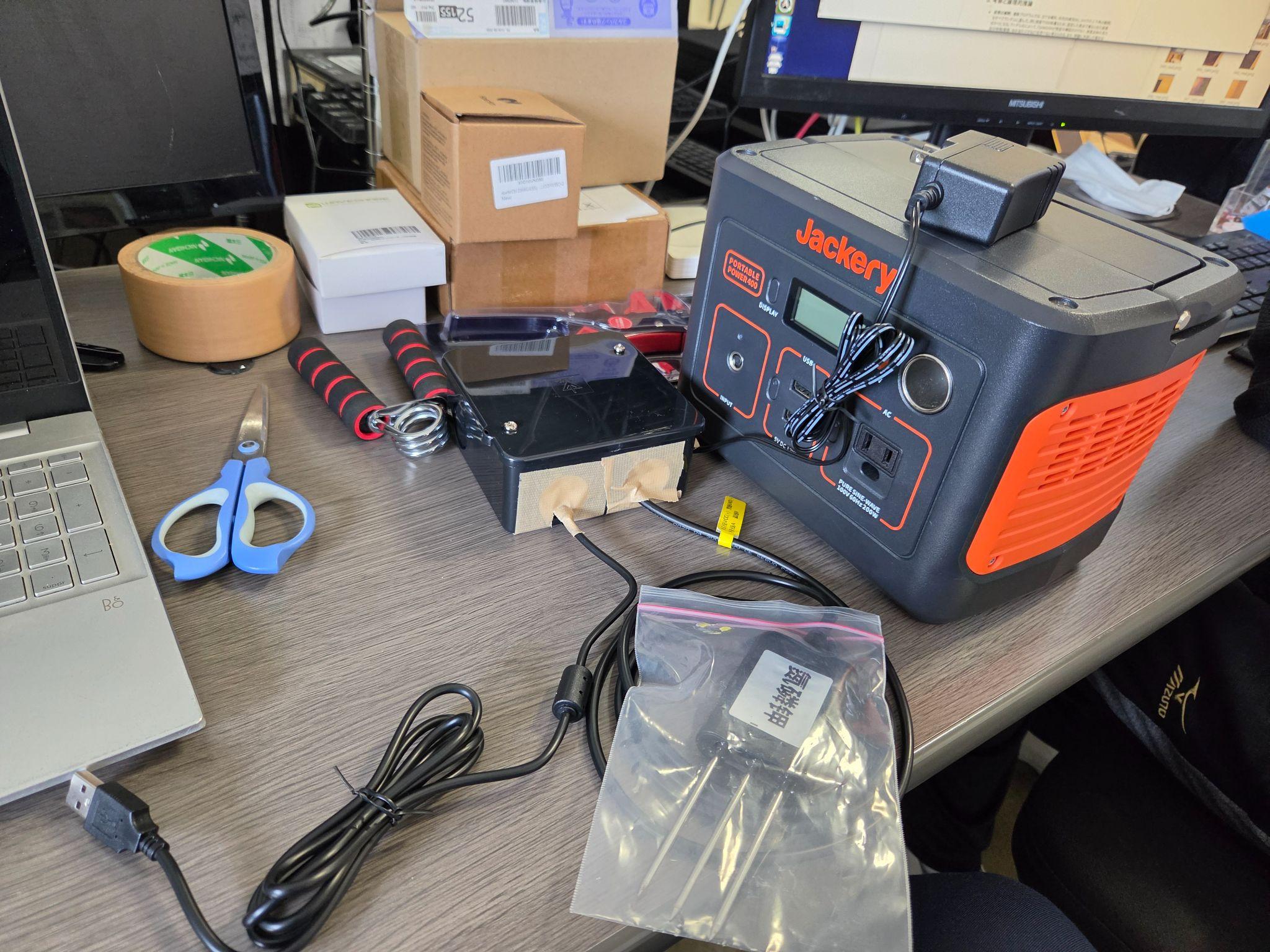}} \\
\midrule
Protective enclosure &
Houses and insulates the internal electrical terminals and converter against dust and mechanical shock during fieldwork. &
Mirai Industry PVK-ANK shallow junction box &
\raisebox{-0.45\height}{\includegraphics[width=2.0cm]{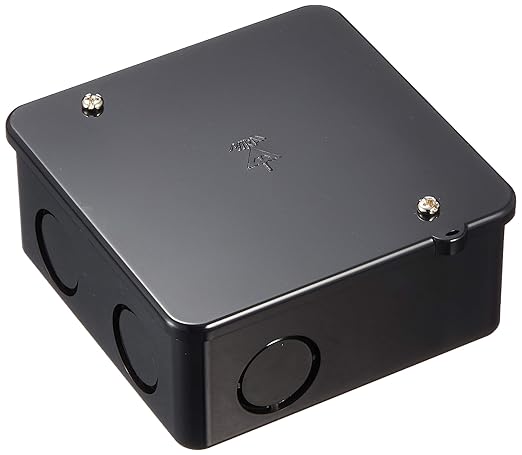}} \\
\midrule
RS485 Modbus soil probe &
Measures in-situ soil macronutrient concentrations (N, P, and K in mg/kg) via metallic sensing prongs. &
RS485 Modbus RTU NPK soil sensor (9600\,baud) &
See Fig.~\ref{fig:assembly} \\
\bottomrule
\end{tabularx}
\end{table}

\begin{figure}[H]
    \centering
    \begin{subfigure}[b]{0.31\textwidth}
        \includegraphics[width=\textwidth]{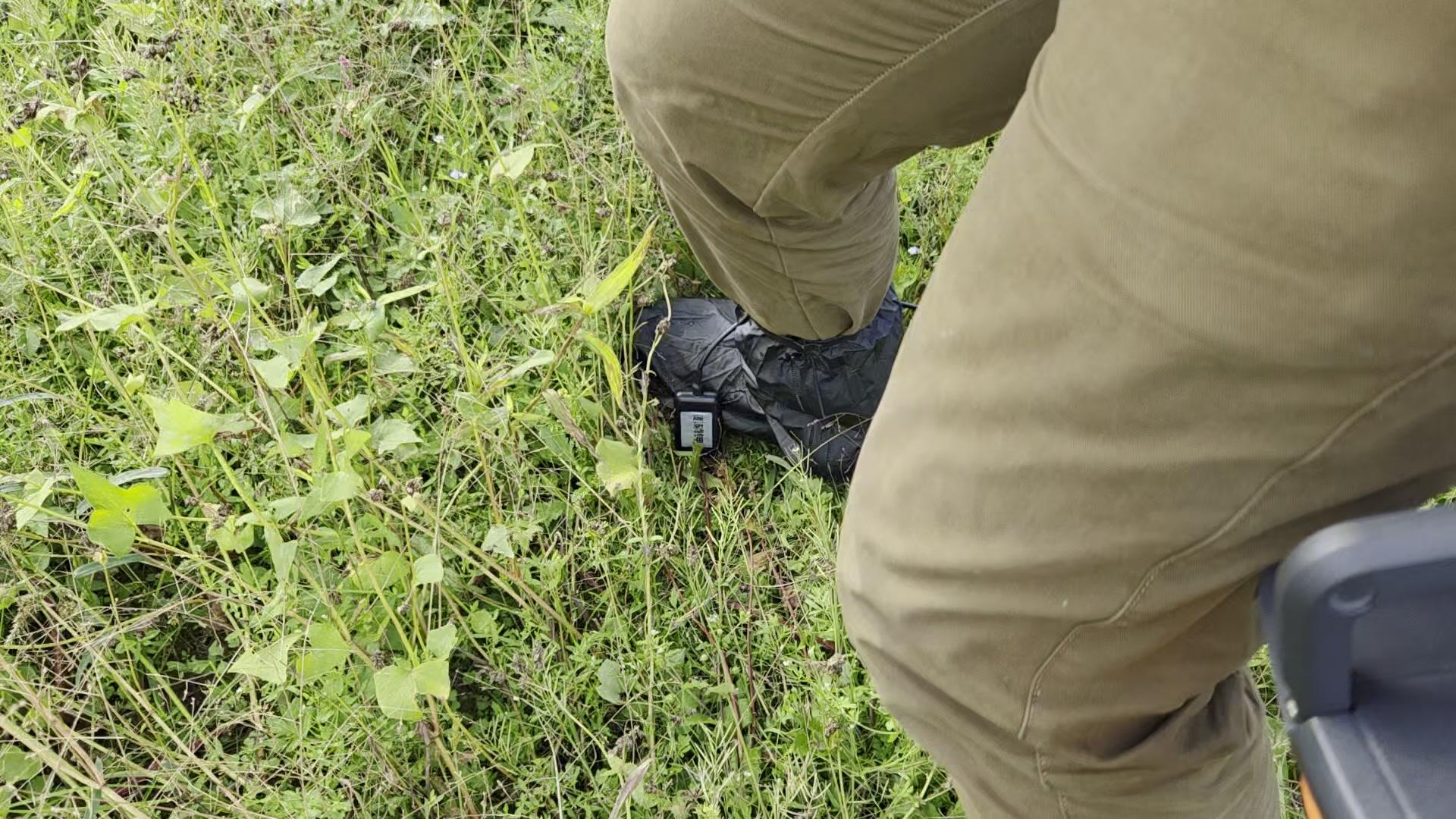}
        \caption{Probe insertion in dense cover}
    \end{subfigure}
    \hfill
    \begin{subfigure}[b]{0.31\textwidth}
        \includegraphics[width=\textwidth]{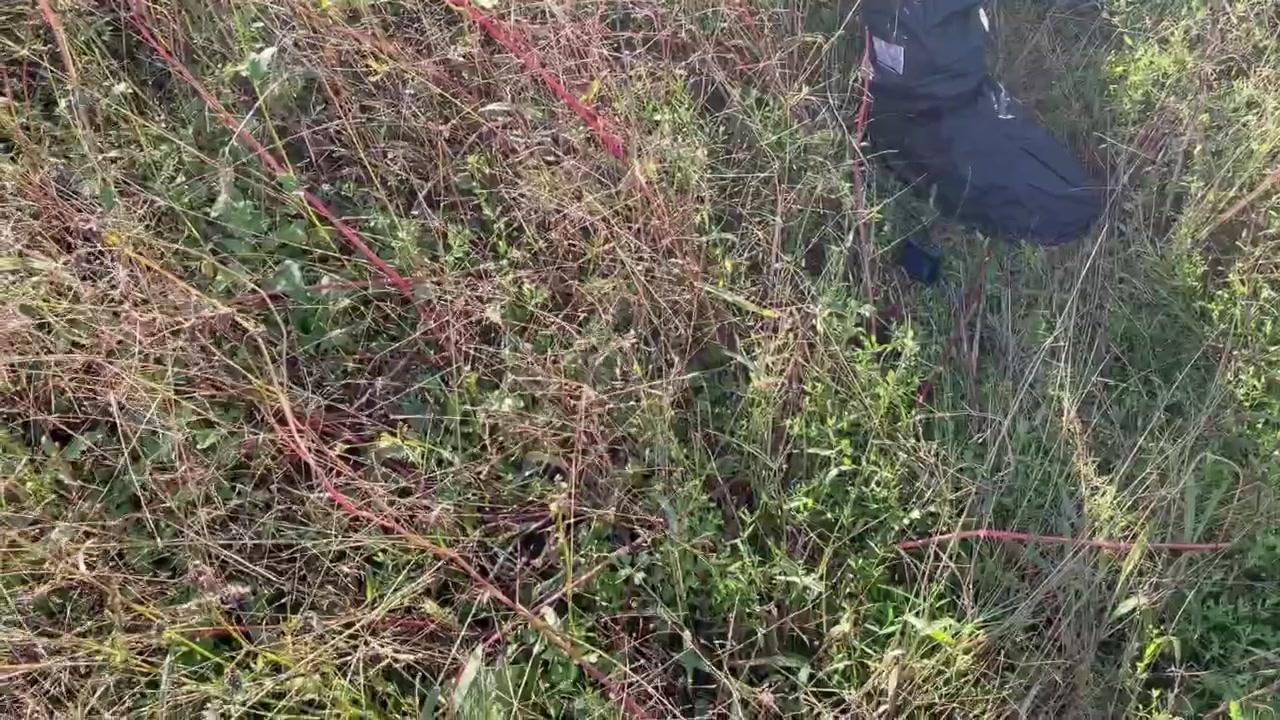}
        \caption{Close-up of sensor body in soil}
    \end{subfigure}
    \hfill
    \begin{subfigure}[b]{0.31\textwidth}
        \includegraphics[width=\textwidth]{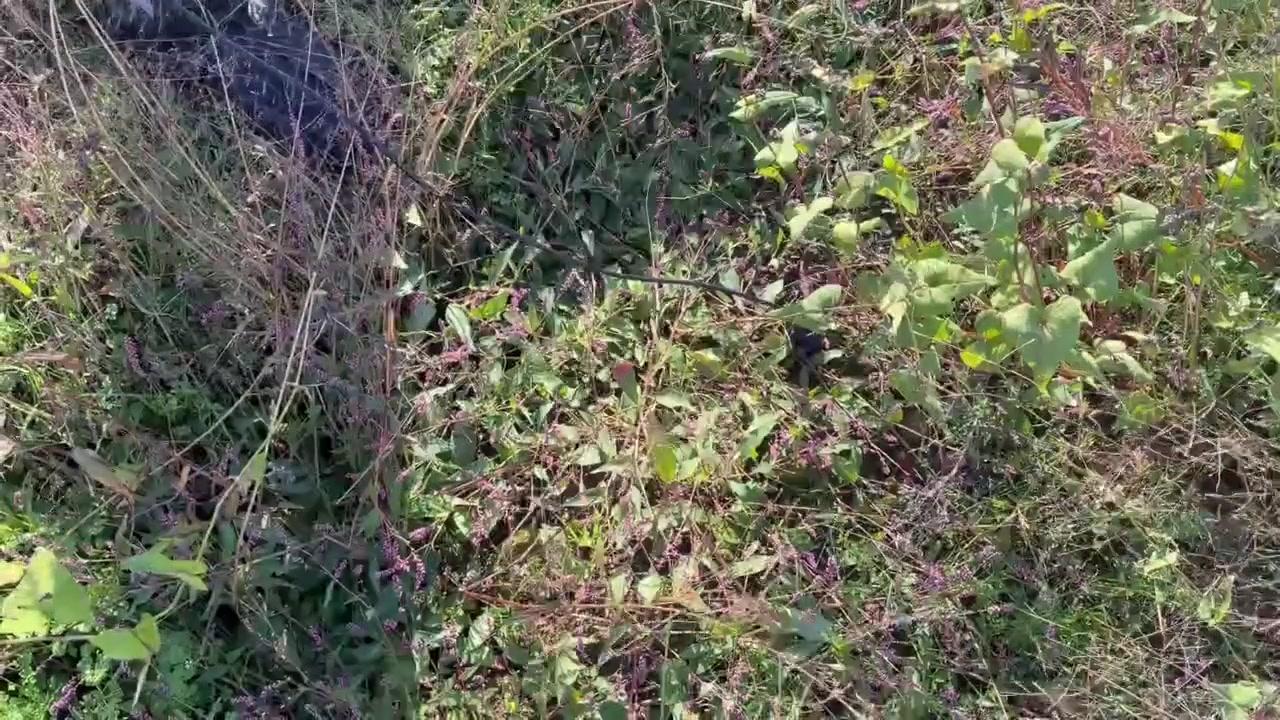}
        \caption{Manual positioning during reading}
    \end{subfigure}
    \caption{Representative frames from the close-up smartphone video documenting probe--soil interaction.}
    \label{fig:closeup}
\end{figure}

\begin{figure}[H]
    \centering
    \begin{subfigure}[b]{0.48\textwidth}
        \includegraphics[width=\textwidth]{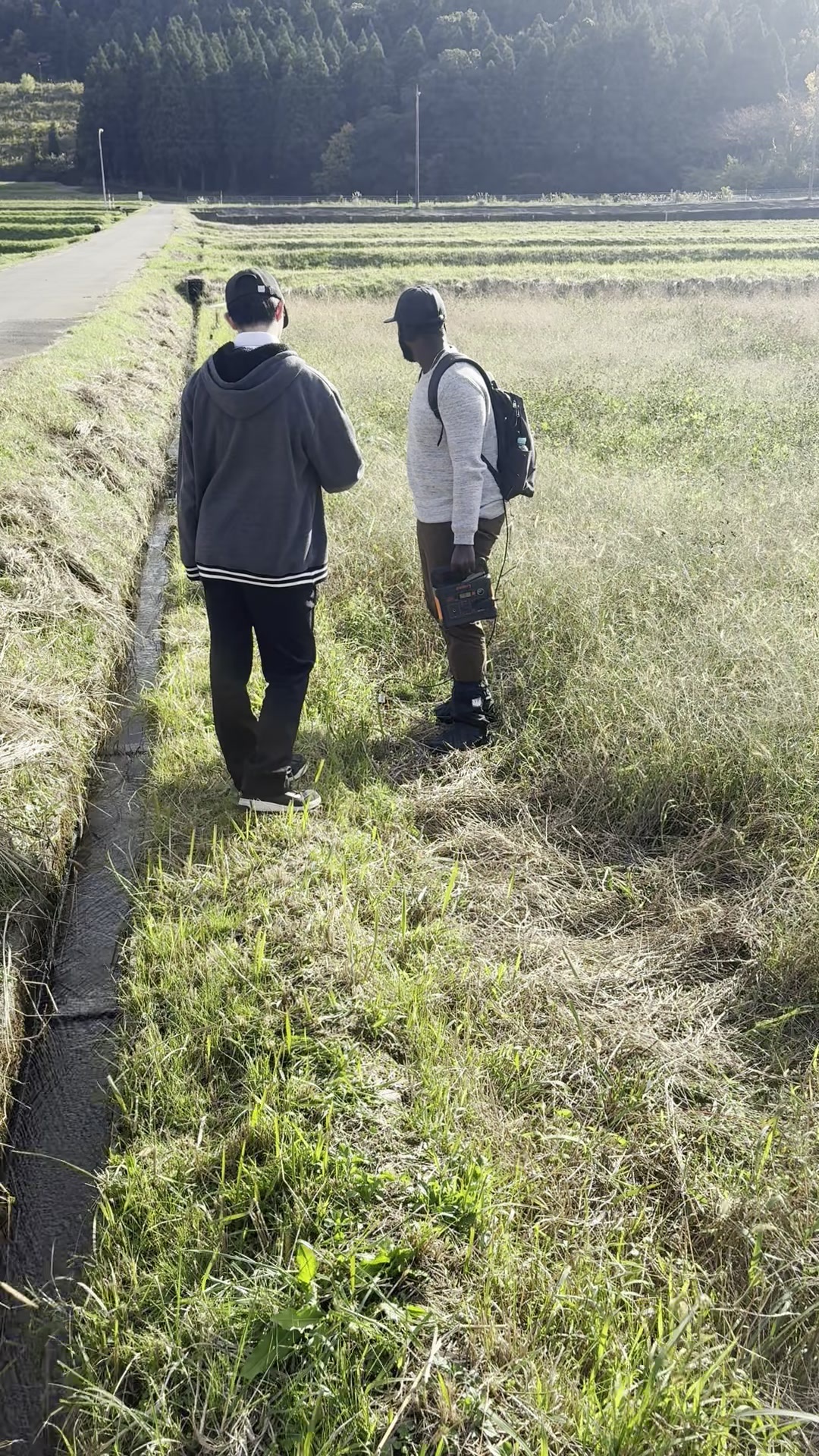}
        \caption{Wide-angle Position 1 (Regions A and B)}
    \end{subfigure}
    \hfill
    \begin{subfigure}[b]{0.48\textwidth}
        \includegraphics[width=\textwidth]{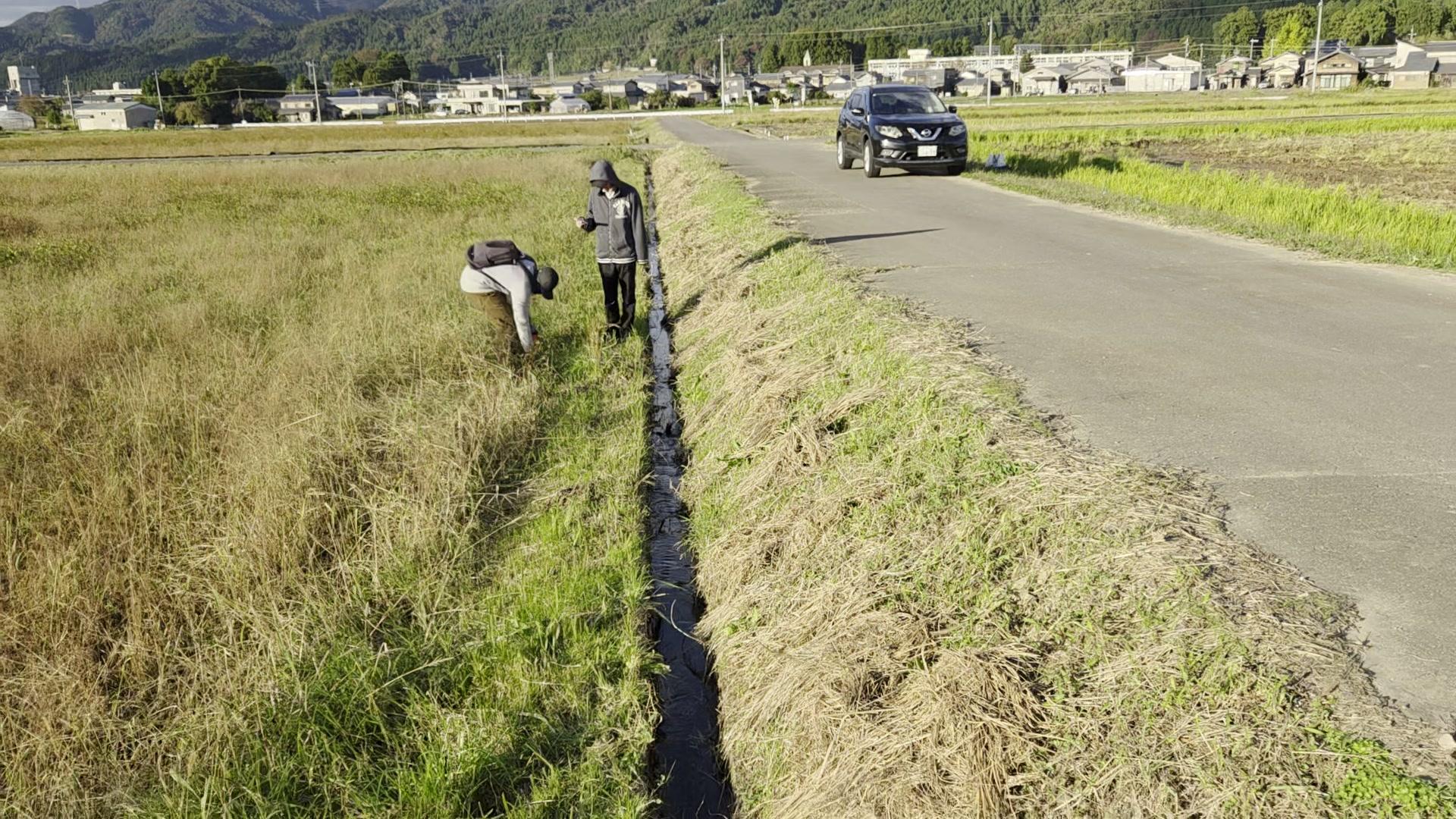}
        \caption{Wide-angle Position 2 (Region C)}
    \end{subfigure}
    \caption{Wide-angle video perspectives capturing operator trajectory and regional context.}
    \label{fig:wideangle}
\end{figure}

Because raw numerical CSV logs alone cannot verify whether the sensing prongs achieved full vertical soil contact or capture surrounding canopy and surface conditions, the prototype system pairs chemical logging with continuous dual-perspective video recording:
\begin{itemize}
    \item \textbf{Close-up Proximal Video Stream:} Recorded continuously at close range using a handheld smartphone (Figure~\ref{fig:closeup}), this proximal perspective serves both to visually audit probe insertion depth, soil contact, and surface vegetation cover at each spot, and to provide the raw video frames from which top-down RGB imagery is extracted for downstream visual-to-chemical modeling.
    \item \textbf{Wide-angle Context Video Stream:} Recorded continuously from stationary end-of-field vantage points (Figure~\ref{fig:wideangle}), this overview perspective documents operator traversal trajectory, boundary crossings, and regional spatial context across the plot.
\end{itemize}
Combining these two synchronized visual streams eliminates single-modality ambiguity and provides an auditable physical record linking every chemical reading to its exact surface appearance and field trajectory.

%% file: a2.tex
\section{Field Operating Protocol (SOP Mapping) and Spatial Partitioning}

\begin{figure}[H]
    \centering
    \includegraphics[width=0.9\textwidth]{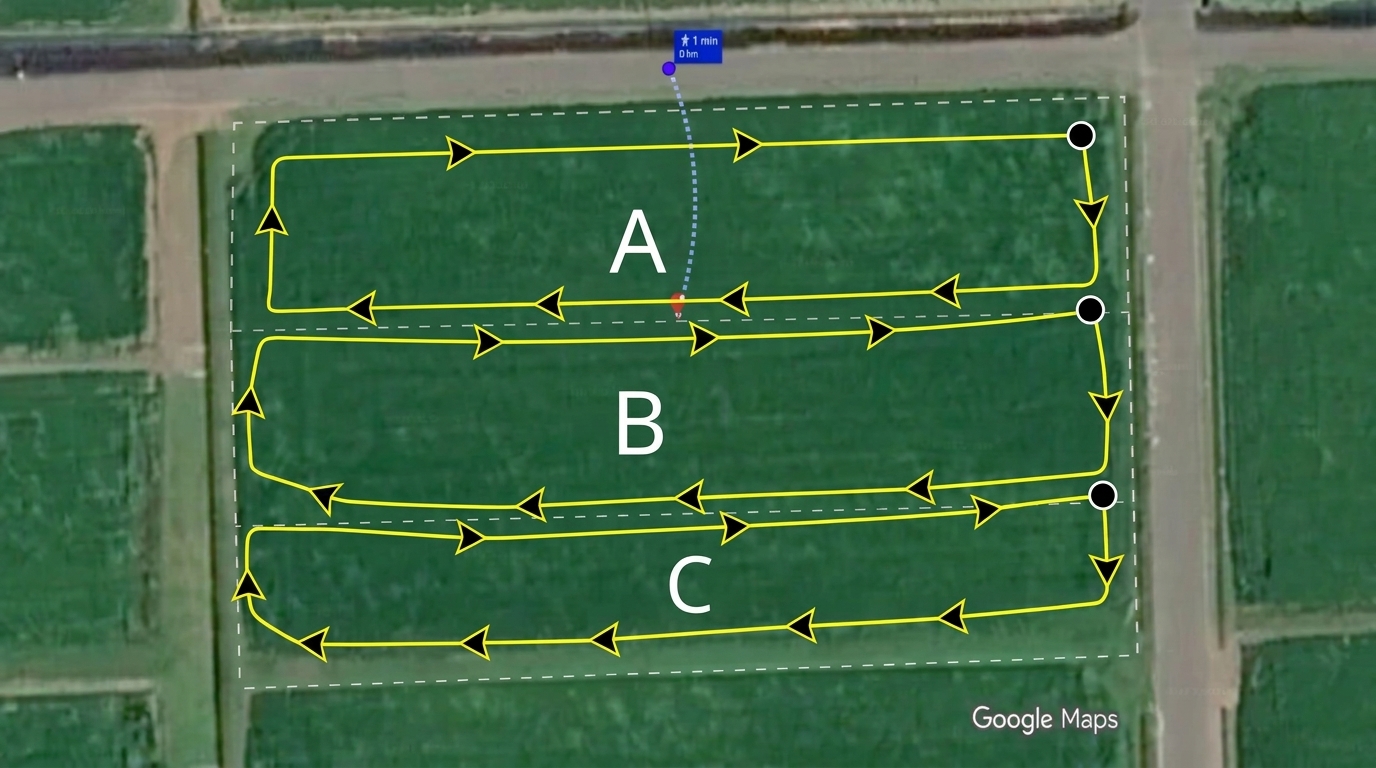}
    \caption{Aerial view of the experimental field showing the boundaries of Regions A, B, and C, starting points (black circles), and the U-shaped traversal path (yellow arrows).}
    \label{fig:layout}
\end{figure}

To ensure repeatable spatial coverage across heterogeneous terrain, the field campaign was conducted at a private rectangular agricultural plot measuring approximately $136\,\text{m} \times 36\,\text{m}$ ($A \approx 4{,}896\,\text{m}^2 \approx 0.49\,\text{ha}$), which was partitioned longitudinally into three contiguous strips labeled \textbf{A}, \textbf{B}, and \textbf{C} (each $\approx 136\,\text{m} \times 12\,\text{m}$, or $\approx 1{,}632\,\text{m}^2$). As illustrated in Figure~\ref{fig:layout} and Table~\ref{tab:team}, a four-person field team executed a continuous U-shaped traversal path within each strip starting from designated corner landmarks: the field supervisor oversaw protocol adherence and boundary transitions; the operator carried the active laptop and sensor box inside a backpack, held the Jackery power station in the left hand, and inserted the RS485 soil probe with the right hand; the close-up video operator continuously recorded proximal smartphone footage of each insertion; and the wide-angle video operator captured stationary overview footage from Position~1 (covering strips \textbf{A} and \textbf{B}) before relocating to Position~2 (covering strip \textbf{C}).

\begin{table}[H]
\centering
\caption{Field team roles and responsibilities during SOP Mapping data acquisition.}
\label{tab:team}
\renewcommand{\arraystretch}{1.35}
\begin{tabularx}{\textwidth}{>{\bfseries}p{4.4cm} Y}
\toprule
\rowcolor{LightTint}
Team member & Assigned responsibility \\
\midrule
Field supervisor & Oversaw protocol adherence, boundary transitions, and operational safety throughout the campaign. \\
Operator (team leader) & Carried the portable acquisition system (backpack laptop/sensor box, left-hand power station, right-hand probe) and performed sequential insertions along the traversal path. \\
Close-up video operator & Continuously recorded proximal smartphone video of the probe and surrounding soil surface to document insertion state and extract top-down RGB frames. \\
Wide-angle video operator & Recorded stationary overview video from one end of the field during traversal of regions \textbf{A} and \textbf{B} (Position~1), then relocated to the opposite end for region \textbf{C} (Position~2). \\
\bottomrule
\end{tabularx}
\end{table}

\begin{figure}[H]
    \centering
    \begin{subfigure}[b]{0.48\textwidth}
        \includegraphics[width=\textwidth]{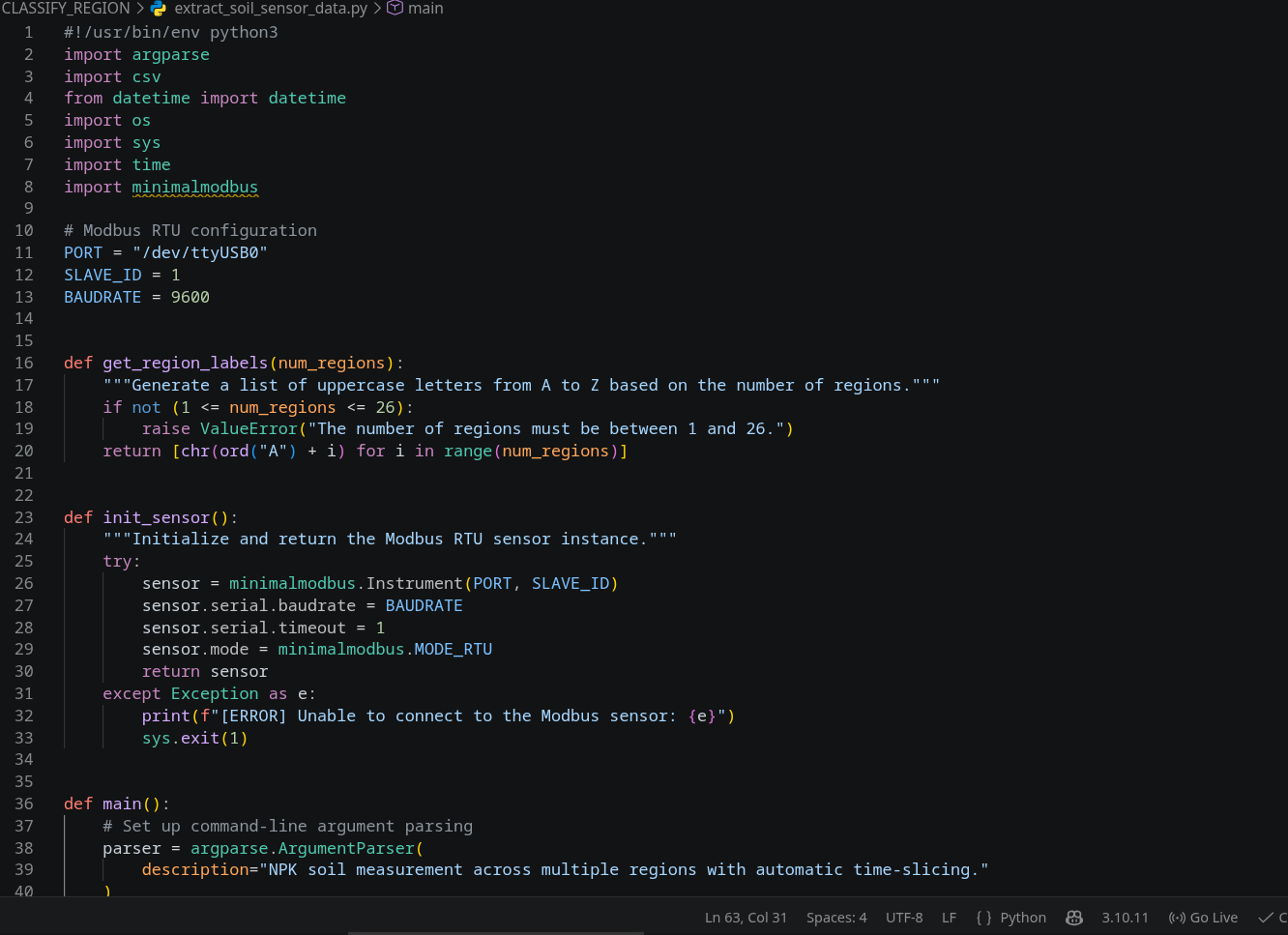}
        \caption{Modbus initialization and region label generator}
    \end{subfigure}
    \hfill
    \begin{subfigure}[b]{0.48\textwidth}
        \includegraphics[width=\textwidth]{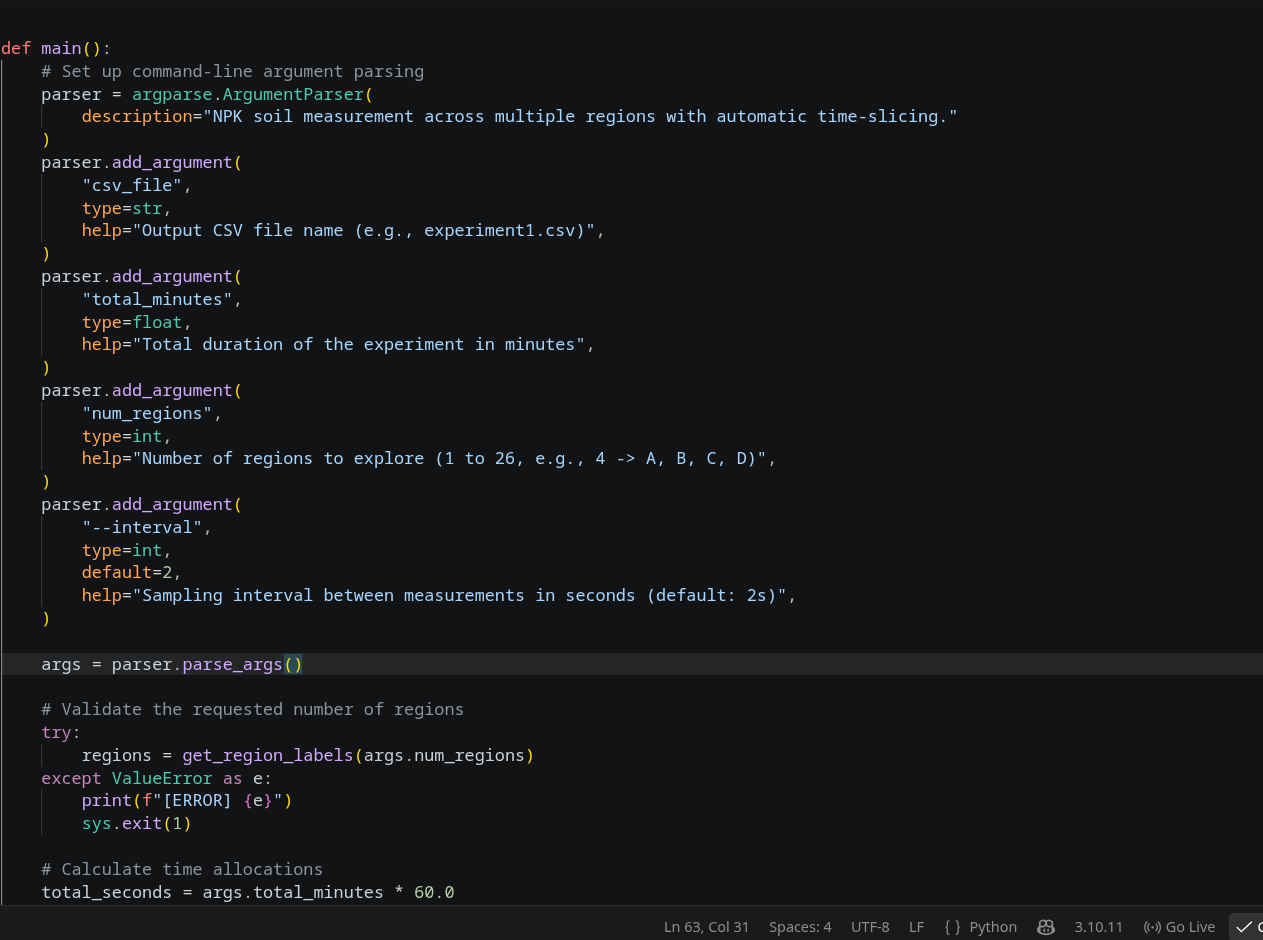}
        \caption{Command-line argument parser setup}
    \end{subfigure}
    
    \vspace{0.35cm}
    
    \begin{subfigure}[b]{0.48\textwidth}
        \includegraphics[width=\textwidth]{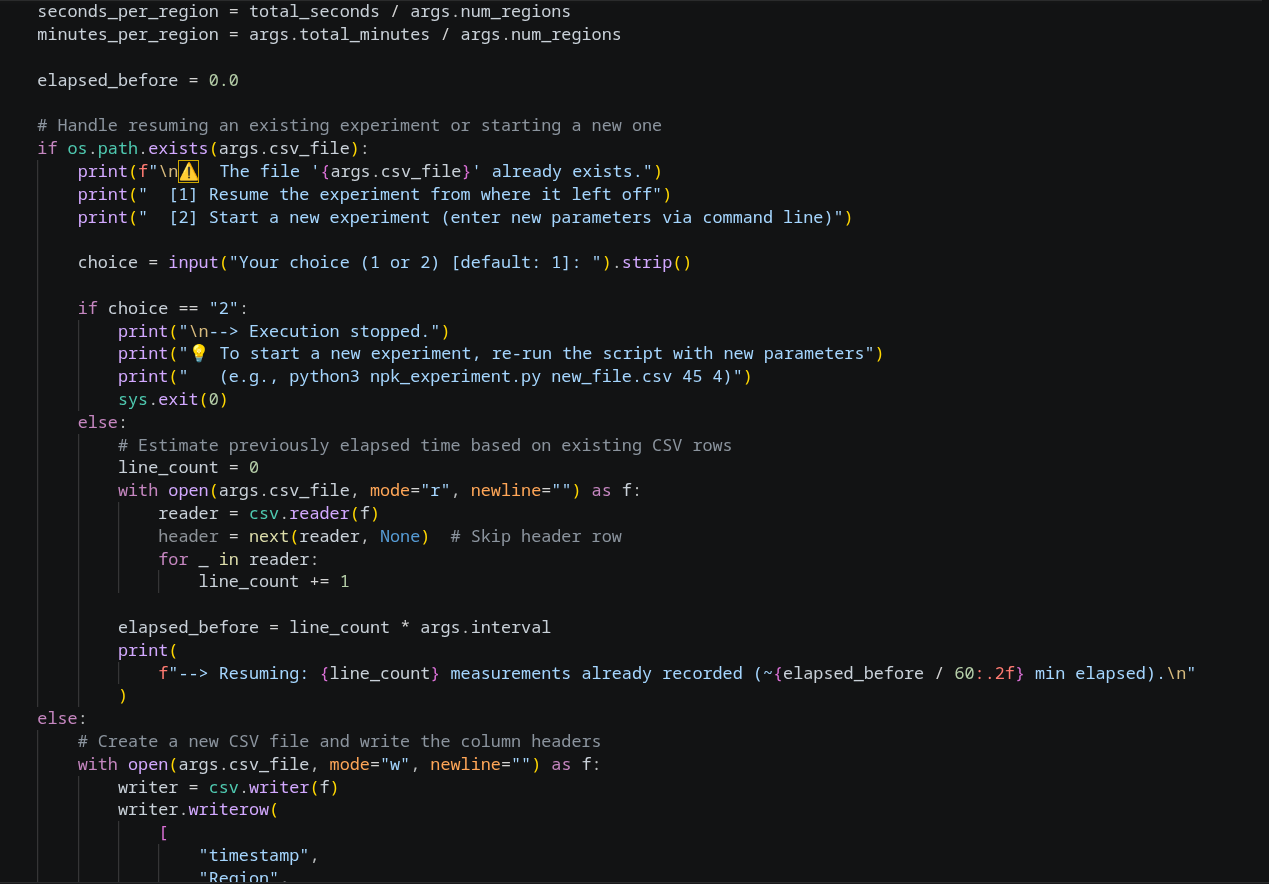}
        \caption{Experiment resume and CSV initialization logic}
    \end{subfigure}
    \hfill
    \begin{subfigure}[b]{0.48\textwidth}
        \includegraphics[width=\textwidth]{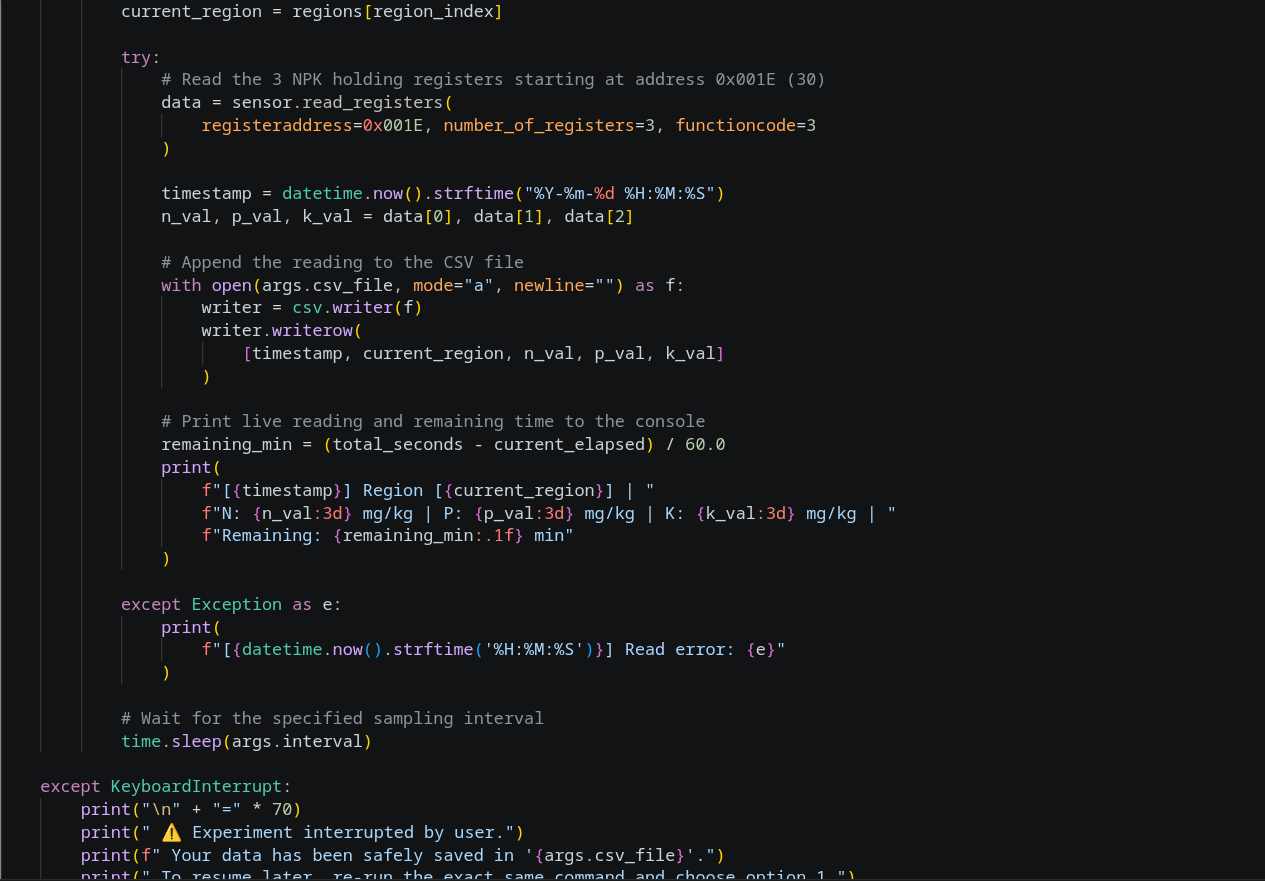}
        \caption{Register polling (\texttt{0x001E}) and CSV logging loop}
    \end{subfigure}
    \caption{Source code implementation of the automated Modbus RTU data-acquisition script.}
    \label{fig:code}
\end{figure}

At each sampling spot along the U-shaped trajectory, the SOP mandates inserting the sensing prongs vertically into the topsoil and holding the probe stationary for approximately \textbf{10\,s} while the Python acquisition script polls the Modbus registers at a nominal 2\,s interval (yielding four to five consecutive records per spot). When operating without live GNSS coordinates (addressing \textbf{RQ4} in Section~3), the acquisition script (Table~\ref{tab:cli} and Figure~\ref{fig:code}) anchors records to spatial strips via equal elapsed-time partitioning, dividing the total scheduled duration $T_{\text{total}}$ across the $N_{\text{regions}}$ requested zones:
\begin{equation}
T_{\text{region}} = \frac{T_{\text{total}}}{N_{\text{regions}}},
\end{equation}
where sequential region identifiers (\texttt{A}, \texttt{B}, \texttt{C}, \dots) are assigned automatically according to elapsed session time (e.g., allocating $10\,\text{min}$ per strip for $T_{\text{total}} = 30\,\text{min}$ and $N_{\text{regions}} = 3$) and cross-checked against physical corner markers and wide-angle video timestamps.

\begin{table}[H]
\centering
\caption{Command-line arguments accepted by the Python data-acquisition script.}
\label{tab:cli}
\renewcommand{\arraystretch}{1.3}
\begin{tabularx}{\textwidth}{>{\bfseries}p{5.2cm} Y}
\toprule
\rowcolor{LightTint}
Command-line parameter & Description \\
\midrule
1. Output CSV filename & Target file path where timestamped measurements are appended (e.g., \texttt{field\_test.csv}). \\
2. Total experiment duration & Total scheduled acquisition window, specified in minutes (e.g., \texttt{60}). \\
3. Number of field regions & Number of sequential spatial zones covered during the session (e.g., \texttt{3} defines regions \texttt{A}, \texttt{B}, and \texttt{C}). \\
4. Sampling interval (\texttt{--interval}) & Optional flag specifying the nominal delay between consecutive register polls, in seconds (default: \texttt{2}\,s). \\
\bottomrule
\end{tabularx}
\end{table}

%% file: a3.tex
\section{Hardware-Grounded Data Cleansing, Cross-Modal Synchronization, and Reproducibility Package}

\begin{figure}[H]
    \centering
    \includegraphics[width=0.92\textwidth]{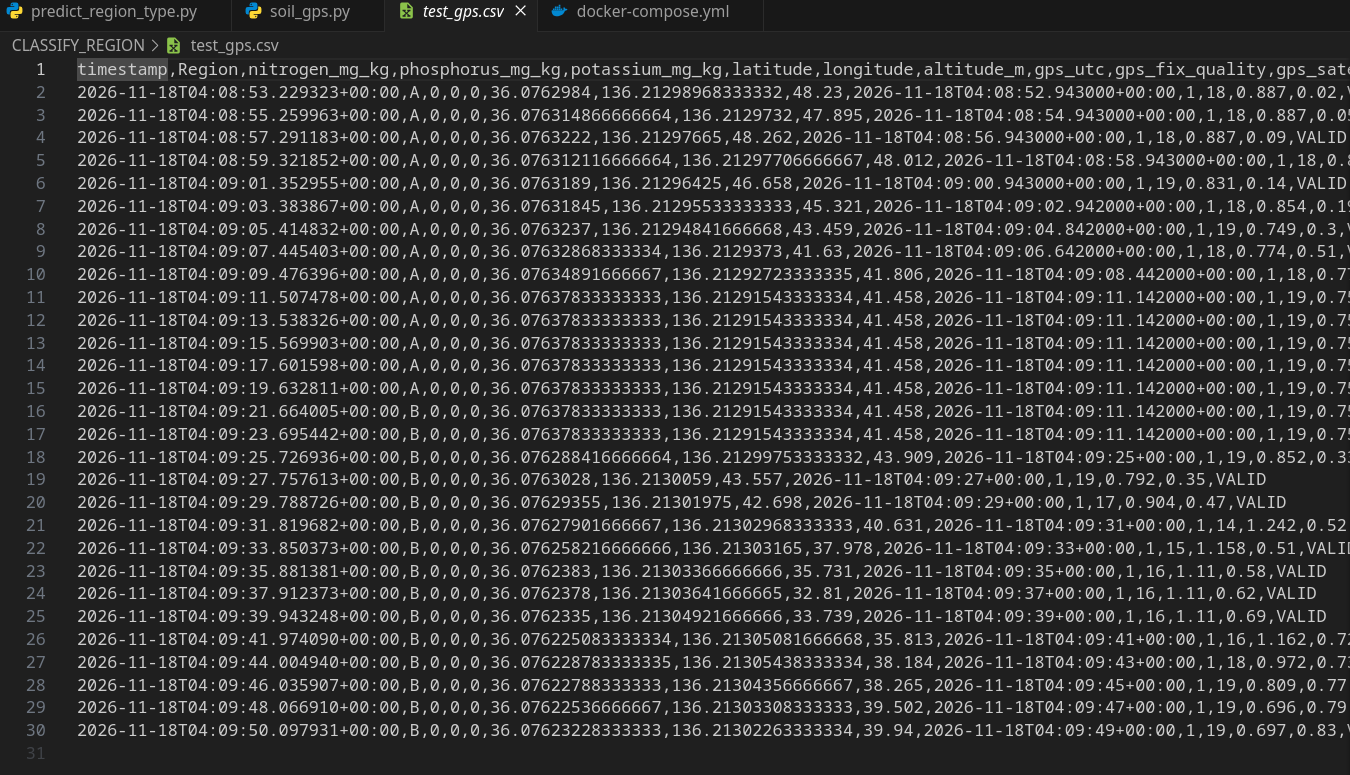}
    \caption{Terminal output during preliminary laboratory testing showing baseline readings in air (\texttt{0\,mg/kg}) and active readings when inserted into soil.}
    \label{fig:prelim}
\end{figure}

\begin{figure}[H]
    \centering
    \includegraphics[width=0.96\textwidth]{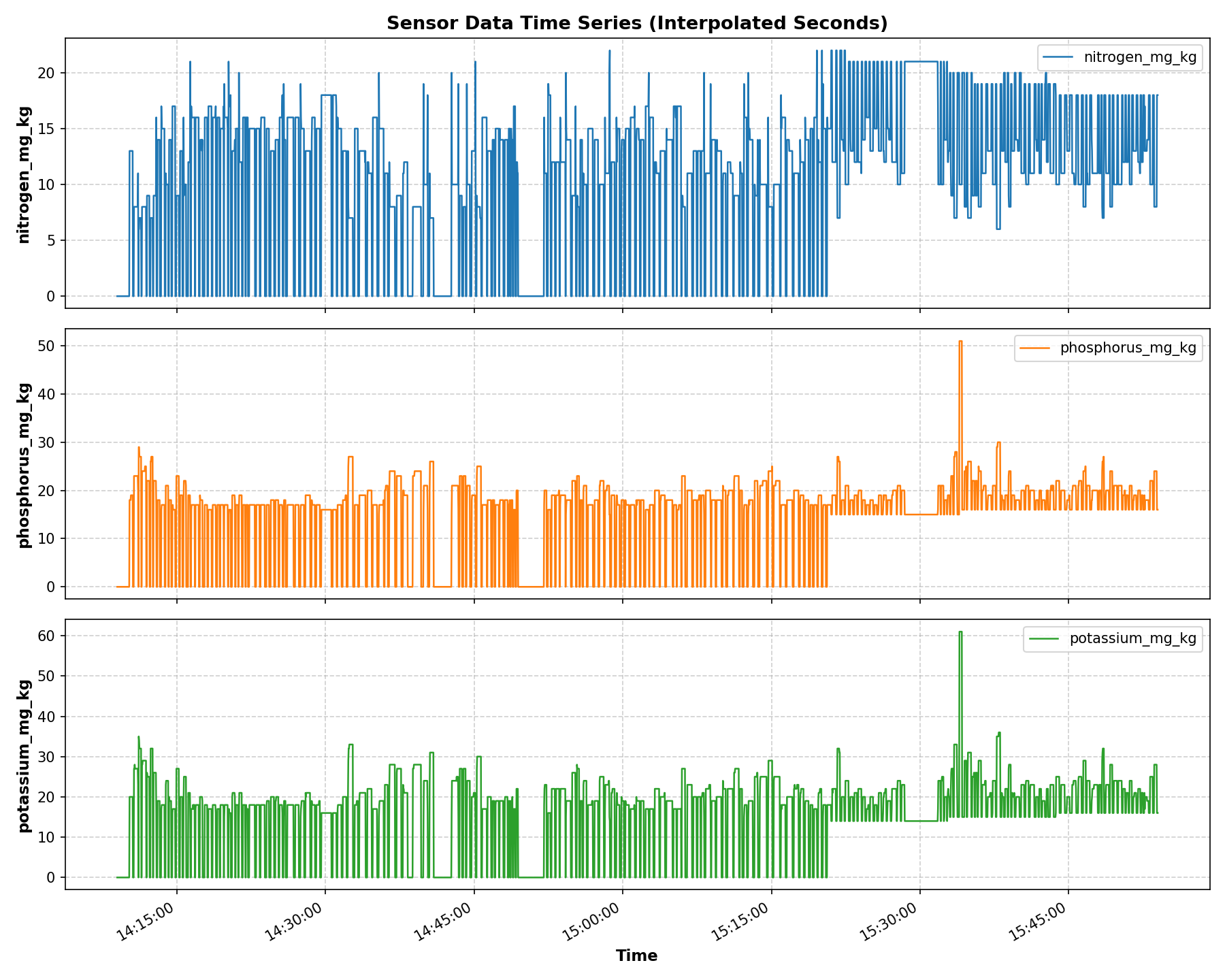}
    \caption{Time-series plot of interpolated valid sensor readings for Nitrogen (\texttt{nitrogen\_mg\_kg}), Phosphorus (\texttt{phosphorus\_mg\_kg}), and Potassium (\texttt{potassium\_mg\_kg}) across regions \textbf{A}, \textbf{B}, and \textbf{C} after non-contact transit filtering.}
    \label{fig:timeseries}
\end{figure}

During continuous field traversal, a key operational challenge is separating active in-soil measurements from non-contact transit intervals while the operator carries the probe between sampling spots. Preliminary laboratory benchtop verification established that the RS485 Modbus probe consistently registers \texttt{0\,mg/kg} for N, P, and K while its sensing prongs remain suspended in ambient air, and jumps immediately to non-zero concentrations upon insertion into soil (Figure~\ref{fig:prelim}). Exploiting this distinct air-versus-soil step response alongside the 10\,s stationary hold rule enables programmatic removal of non-contact transit records without labor-intensive manual annotation, yielding clean spot-level time series across strips \textbf{A}, \textbf{B}, and \textbf{C} (Figure~\ref{fig:timeseries}). Note that the \texttt{0\,mg/kg} air reading reflects the operational behavior of this specific Modbus probe under non-contact conditions and serves strictly as a contact-state indicator rather than a universal zero-point chemical standard.

To align top-down RGB frames $\{I_1, \dots, I_N\}$ extracted from the close-up smartphone video (with acquisition timestamps $t_i$) against discrete in-soil spot measurement intervals $[t_j^{\mathrm{start}}, t_j^{\mathrm{end}}]$ (temporal midpoint $m_j = \frac{1}{2}(t_j^{\mathrm{start}} + t_j^{\mathrm{end}})$ and spot-averaged nutrient vector $\bar{\mathbf{y}}_j \in \mathbb{R}^3$), each image is paired with a reference label $\mathbf{y}_i = \bar{\mathbf{y}}_{j^*(i)}$ via interval containment with nearest-midpoint fallback:
\begin{equation}
j^*(i) = 
\begin{cases} 
\arg\min_{j \in \mathcal{S}_i} \lvert t_i - m_j \rvert & \text{if } \mathcal{S}_i = \{ j \mid t_j^{\mathrm{start}} \le t_i \le t_j^{\mathrm{end}} \} \neq \emptyset, \\[6pt]
\arg\min_{j} \lvert t_i - m_j \rvert & \text{otherwise, subject to } \Delta t_i = \min_{j} \lvert t_i - m_j \rvert \le \tau_{\max},
\end{cases}
\end{equation}
where $\tau_{\max} = 84\text{ s}$ enforces a strict maximum cross-modal clock tolerance (discarding frames exceeding $\tau_{\max}$ to prevent spatial label mismatch). Applying this synchronization rule produced $N = 242$ validated image--nutrient pairs ($155$ matched directly within active measurement intervals and $87$ matched within $\tau_{\max}$). To ensure full experimental transparency and auditability for downstream benchmarking, the complete archival reproducibility package comprises:
\begin{itemize}
    \item \textbf{Raw and Processed Sensor Logs:} The unmodified CSV output generated directly by the Modbus acquisition script alongside cleaned spot-level CSV tables and exact transit-exclusion thresholds.
    \item \textbf{Acquisition and Synchronization Software:} Version-controlled Python scripts (\texttt{extract\_soil\_sensor\_data.py} / \texttt{npk\_experiment.py}), CLI invocation parameters, and cross-modal timestamp alignment code.
    \item \textbf{Multimodal Visual Records:} Original close-up and wide-angle video files (with original filenames, frame rates, and creation timestamps) together with the $N = 242$ synchronized top-down RGB images.
    \item \textbf{Hardware, Spatial, and Operational Metadata:} Component bill of materials, RS485 Modbus framing specifications (\texttt{9600\,baud}, 8N1, Slave ID~1, holding register \texttt{0x001E}), 12\,V DC power wiring schematics, annotated aerial field maps (strips \textbf{A}--\textbf{C} and U-shaped paths), and field operational notes.
\end{itemize}

%% file: b1.tex
\section{Sanity-Check Objective and Lightweight Baseline Pipeline}

Before utilizing the prototype dataset ($N = 242$) constructed via SOP Mapping (Supplementary Material~A) as a benchmark for heavy foundation models (e.g., SAM~2.1 and DINOv3), it is necessary to perform a diagnostic \textbf{sanity check}. Specifically, we aim to verify that the synchronized top-down RGB frames and spot-level in-situ chemical measurements retain a statistically non-trivial agronomic signal—rather than being dominated by cross-modal clock-offset noise, probe stabilization artifacts, or outdoor visual clutter. To keep this diagnostic transparent and interpretable under a limited sample size ($N < 300$), we deliberately employ a lightweight, rule-based foreground masking and handcrafted feature pipeline paired with standard linear, regularized, and ensemble regressors.

To suppress obvious non-target outdoor artifacts—namely deep cast shadows, operator footwear, yellow equipment enclosures, and specular glare—we construct a binary valid foreground mask $M(u, v) \in \{0, 1\}$ in OpenCV HSV space ($H \in [0, 180]$, $S, V \in [0, 255]$) at each pixel $(u, v)$:
\begin{equation}
M(u, v) = \mathbb{I}\bigl(V(u, v) > 45\bigr) \cdot \mathbb{I}\bigl(V(u, v) < 245\bigr) \cdot \Bigl[1 - \mathbb{I}\bigl(14 \le H(u, v) \le 34 \;\wedge\; S(u, v) > 120\bigr)\Bigr],
\end{equation}
reverting to the full image domain if $\sum_{u,v} M(u, v) < 100$. From each image $I_i$ and its mask $M_i$, we extract a 23-dimensional handcrafted descriptor $\mathbf{x}_i \in \mathbb{R}^{23}$ comprising three modular subsets:
\begin{itemize}
    \item \textbf{Global Color Statistics (\texttt{color\_hsv}, 8 dims):} Channel-wise spatial means of RGB and normalized HSV, plus the mean and standard deviation of the Excess Green index ($\mathrm{ExG} = 2G - R - B$).
    \item \textbf{Spatial Texture Descriptors (\texttt{glcm}, 5 dims):} Five Haralick Gray-Level Co-occurrence Matrix (GLCM) properties (contrast, dissimilarity, homogeneity, energy, and correlation) averaged over distances $d \in \{1, 3\}$ and angles $\theta \in \{0, \pi/2\}$.
    \item \textbf{Masked Vegetation and Perceptual Indices (\texttt{veg\_masked}, 10 dims):} Computed strictly within the valid foreground domain $\Omega_i = \{ (u, v) \mid M_i(u, v) = 1 \}$, including valid pixel fraction, green canopy fraction ($\mathrm{ExG} > 0.05$), masked mean and standard deviation of $\mathrm{ExG}$, masked spatial means of $\mathrm{VARI} = \mathrm{clip}\bigl((G - R)/(G + R - B + \epsilon), -1, 1\bigr)$, $\mathrm{GLI} = (2G - R - B)/(2G + R + B + \epsilon)$, and CIE $L^*a^*b^*$ chromaticity channels ($\mu_{a^*}, \mu_{b^*}$), alongside masked GLCM contrast and homogeneity.
\end{itemize}
To evaluate whether these cues predict spot-level concentrations $\mathbf{y}_i = (y_i^{\mathrm{N}}, y_i^{\mathrm{P}}, y_i^{\mathrm{K}})^\top$ without data leakage, we benchmark five configurations—\texttt{real01\_color} (Partial Least Squares, $\mathrm{PLS}$ with $L = 5$ components on \texttt{color\_hsv}), \texttt{real02\_color\_glcm} ($\mathrm{PLS}$ on \texttt{color\_hsv}+\texttt{glcm}), \texttt{real04\_all\_handcrafted} ($\mathrm{PLS}$ on all 23 features), \texttt{real05\_ridge} ($L_2$-regularized linear regression, \texttt{RidgeCV} with $\alpha \in [10^{-2}, 10^4]$ on all 23 features), and \texttt{real06\_rf} (Random Forest, $\mathrm{RF}$ with $T = 50$ trees and max depth 6 on all 23 features)—using \textbf{Repeated 24-Fold Stratified Cross-Validation} ($20$ repeats, $480$ folds in total, stratified across $5$ target quantile bins), performing all feature scaling and hyperparameter tuning strictly inside each training fold.

%% file: b2.tex
\section{Sanity-Check Results: Verification of Non-Trivial Agronomic Signal}

\begin{figure}[ht]\centering
\includegraphics[width=\textwidth]{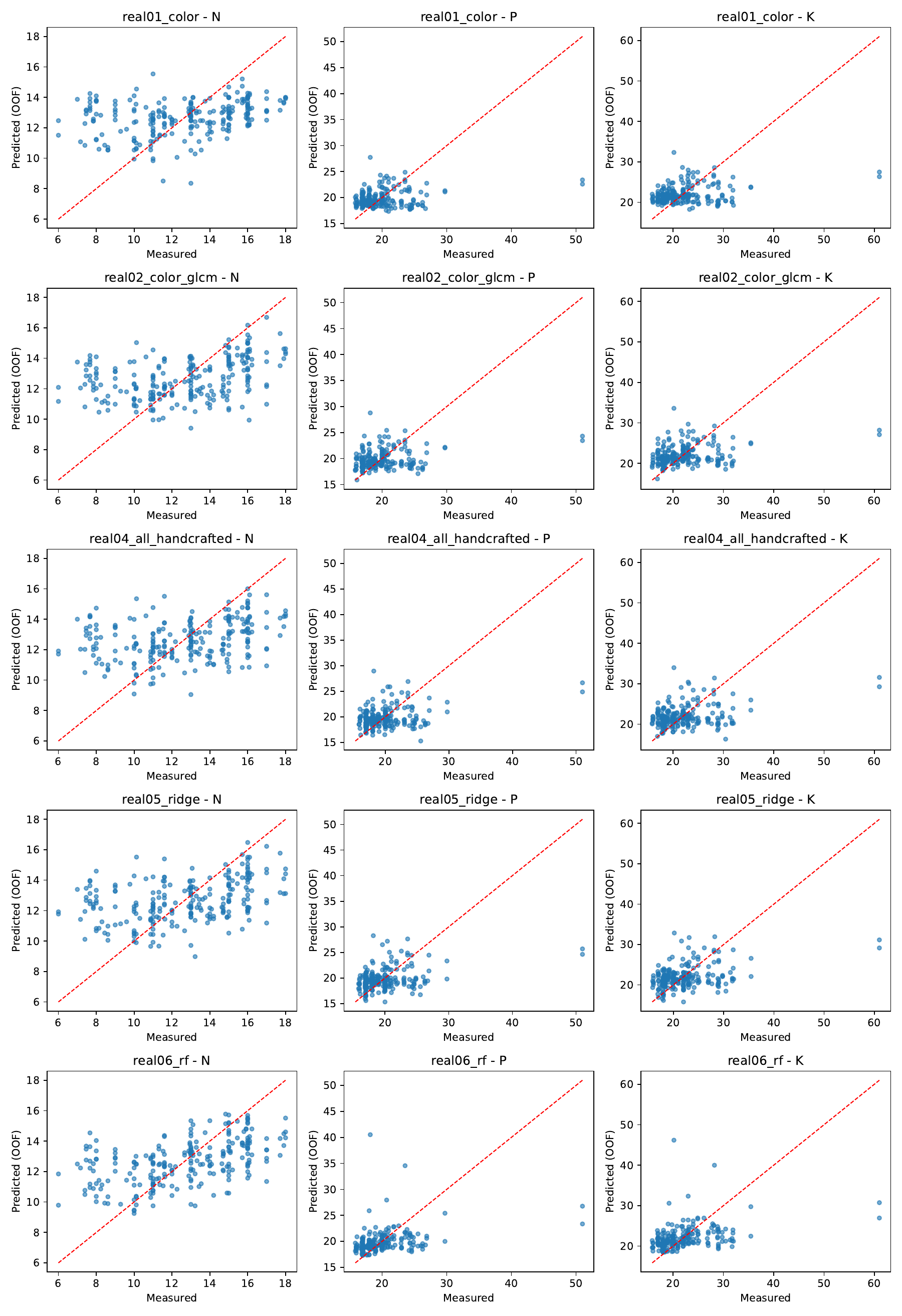}
\caption{Measured vs.\ Out-of-Fold Predicted Values across baseline feature sets and regression models, illustrating non-trivial signal recovery for Nitrogen alongside strong mean-regression compression at high-concentration outliers.}
\label{fig:scatter}
\end{figure}

Table~\ref{tab:cv_results} and Figure~\ref{fig:scatter} summarize the out-of-fold Root Mean Squared Error ($\mathrm{RMSE}$) and coefficient of determination ($\mathrm{OOF}\ R^2$) across all five baseline configurations. Crucially for our diagnostic sanity check, the dataset exhibits statistically non-trivial visual-to-chemical predictability rather than random noise—most clearly for Nitrogen ($\mathrm{N}$), where the non-linear Random Forest baseline (\texttt{real06\_rf}) reduces $\mathrm{RMSE}$ to $2.47\ \mathrm{mg/kg}$ and reaches $\mathrm{OOF}\ R^2 = 0.24$ (nearly double the linear color-only baseline of $0.08$). This positive out-of-fold signal confirms that the cross-modal timestamp synchronization ($\tau_{\max} = 84\text{ s}$) and non-contact transit filtering developed in Supplementary Material~A successfully aligned proximal canopy and surface imagery with genuine in-situ soil nutrient variations.

\begin{table}[ht]\centering
\caption{Sanity-Check Repeated 24-Fold CV Results ($\mathrm{RMSE}$ in $\mathrm{mg/kg}$ and $\mathrm{OOF}\ R^2$ across $20$ repeats).}
\label{tab:cv_results}
\begin{tabular}{lccc}
\toprule
Experiment & N $\mathrm{RMSE}$ ($R^2$) & P $\mathrm{RMSE}$ ($R^2$) & K $\mathrm{RMSE}$ ($R^2$) \\
\midrule
\texttt{real01\_color} & 2.73 (0.08) & 3.44 (0.05) & 4.81 (0.07) \\
\texttt{real02\_color\_glcm} & 2.65 (0.13) & 3.41 (0.07) & 4.80 (0.07) \\
\texttt{real04\_all\_handcrafted} & 2.74 (0.07) & 3.44 (0.09) & 4.80 (0.10) \\
\texttt{real05\_ridge} & 2.64 (0.14) & \textbf{3.40 (0.10)} & \textbf{4.70 (0.13)} \\
\texttt{real06\_rf} & \textbf{2.47 (0.24)} & 3.33 (0.05) & 4.63 (0.10) \\
\bottomrule
\end{tabular}
\end{table}

Furthermore, incremental feature comparisons verify that the lightweight artifact-masking and feature-extraction pipeline behaves consistently with domain expectations: adding spatial GLCM texture descriptors (\texttt{real02\_color\_glcm}) improves Nitrogen $\mathrm{OOF}\ R^2$ from $0.08$ to $0.13$, while incorporating artifact-masked vegetation indices (\texttt{real04\_all\_handcrafted}) raises linear predictability for Phosphorus ($0.05 \rightarrow 0.09$) and Potassium ($0.07 \rightarrow 0.10$). Applying $L_2$ regularization (\texttt{real05\_ridge}) stabilizes variance inflation across the 23 correlated predictors, yielding the best linear baselines for Phosphorus ($\mathrm{OOF}\ R^2 = 0.10$, $\mathrm{RMSE} = 3.40\ \mathrm{mg/kg}$) and Potassium ($\mathrm{OOF}\ R^2 = 0.13$, $\mathrm{RMSE} = 4.70\ \mathrm{mg/kg}$). Together, these diagnostic patterns confirm that the dataset and leak-aware validation harness provide a reliable, well-behaved baseline testbed.

%% file: b3.tex
\section{Open Challenges and Substantial Research Headroom}

While the sanity-check results in Section~B.2 confirm that the prototype dataset captures valid agronomic signals, they simultaneously demonstrate that proximal soil mapping remains largely unsolved, with substantial headroom for methodological innovation. Most notably, predictive explanatory power remains severely limited for Phosphorus ($\mathrm{OOF}\ R^2 = 0.10$) and Potassium ($\mathrm{OOF}\ R^2 = 0.13$), and across all three macronutrients, out-of-fold predictions suffer from pronounced regression-toward-the-mean compression at extreme high-concentration outliers (particularly for $\mathrm{P} > 45\ \mathrm{mg/kg}$ and $\mathrm{K} > 55\ \mathrm{mg/kg}$ in Figure~\ref{fig:scatter}). These error patterns underscore \textbf{RQ1} (Section~3): subsurface chemical dynamics of Phosphorus and Potassium exhibit weak direct coupling with visible surface reflectance under small sample sizes ($N = 242$), requiring multi-modal fusion or uncertainty-aware abstention rather than naive point regression.

Second, inspection of failure cases highlights the bottleneck of heuristic visual preprocessing (\textbf{RQ2} and \textbf{Module~2}). Although rule-based thresholding in OpenCV HSV space successfully removes high-contrast artificial confounders such as yellow equipment boxes and deep cast shadows, pixel-wise color rules cannot semantically disentangle exposed soil clods from senescent crop residues, dry litter, or heterogeneous surface crusting. Overcoming this perceptual bottleneck requires replacing heuristic color masks with foundation semantic segmentation models (such as SAM~2.1) to construct category-specific masks (bare soil vs.\ living canopy vs.\ residue) and evaluating self-supervised visual representations (such as DINOv3) against our 23-dimensional handcrafted baseline.

Finally, several foundational components of the grand architecture envisioned in Section~2 remain completely open for future investigation (\textbf{RQ3}, \textbf{RQ4}, and \textbf{Modules~3--5}):
\begin{itemize}
    \item \textbf{Spatially Independent Generalization:} Our diagnostic evaluation relied on Repeated 24-Fold Stratified Cross-Validation within a single field campaign ($N = 242$); transitioning to spatially and temporally grouped validation across held-out plots, soil types, and survey dates is essential to eliminate spatial autocorrelation leakage.
    \item \textbf{Predictive Uncertainty and Abstention:} Developing calibrated uncertainty estimators capable of flagging out-of-distribution soil conditions or weak optical observability remains unaddressed.
    \item \textbf{Geospatial Interpolation and Decision Support:} Integrating GNSS-coupled or SOP-anchored predictions into continuous geostatistical maps (e.g., via kriging) and coupling them with crop-physiological and economic models for variable-rate management represent major downstream research frontiers.
\end{itemize}